\documentclass[conference]{IEEEtran}
\ifCLASSINFOpdf
\else
\fi
\usepackage{cite}
\usepackage{graphicx} 
\usepackage{epstopdf}
\usepackage{color}
\usepackage{caption}
\usepackage{amsmath}
\usepackage{url}
\usepackage{hyperref}
\newcommand*{\affaddr}[1]{#1} 
\newcommand*{\affmark}[1][*]{\textsuperscript{#1}}

\begin{document}
%
\title{A post-processing method to improve the white matter hyperintensity segmentation accuracy for randomly-initialized U-net}

\author{
Yue Zhang\affmark[1, 2], Wanli Chen\affmark[1], Yifan Chen\affmark[1, 3, ]\IEEEauthorrefmark{2}, and Xiaoying Tang\affmark[1, ]\IEEEauthorrefmark{1}\\
\affaddr{\affmark[1]\begin{normalsize}Department of Electrical and Electronic Engineering, Southern University of Science and Technology, Shenzhen, China \end{normalsize}}\\
\affaddr{\affmark[2]\begin{normalsize}Department of Electrical and Electronic Engineering, The University of Hong Kong, Hong Kong SAR, China} \end{normalsize}\\
\affaddr{\affmark[3]\begin{normalsize}Faculty of Science and Engineering, The University of Waikato, Hamilton, New Zealand} \end{normalsize} \\
\affaddr{\IEEEauthorrefmark{2}yifanc@waikato.an.nz, \IEEEauthorrefmark{1}tangxy@sustc.edu.cn }\\
}


\maketitle

\begin{abstract}
    White matter hyperintensity (WMH) is commonly found in elder individuals and appears to be associated with brain diseases. U-net is a convolutional network that has been widely used for biomedical image segmentation. Recently, U-net has been successfully applied to WMH segmentation. Random initialization is usally used to initialize the model weights in the U-net. However, the model may coverage to different local optima with different randomly initialized weights. We find a combination of thresholding and averaging the outputs of U-nets with different random initializations can largely improve the WMH segmentation accuracy. Based on this observation, we propose a post-processing technique concerning the way how averaging and thresholding are conducted. Specifically, we first transfer the score maps from three U-nets to binary masks via thresholding and then average those binary masks to obtain the final WMH segmentation. Both quantitative analysis (via the Dice similarity coefficient) and qualitative analysis (via visual examinations) reveal the superior performance of the proposed method. This post-processing technique is independent of the model used. As such, it can also be applied to situations where other deep learning models are employed, especially when random initialization is adopted and pre-training is unavailable.
\end{abstract}


%
\IEEEpeerreviewmaketitle


\section{Introduction}
White matter hyperintensity (WMH) is commonly found in elderly people and appears to be associated with various brain diseases, such as ischemic stroke, Alzheimer¡¯s disease and multiple sclerosis \cite{1}. The location and size of WMH regions are important biomarkers of these diseases. Manual delineation by experienced neuroradiologists is a reliable way to segment the WMH regions but it is laborious and time-consuming and has high intra- as well as inter-rater variability. As such, automatically segmenting the WMH regions is of great importance, especially in the context of large-scale neuroimaging studies. However, automatic WMH segmentation is challenging, especially for the small WMH lesions \cite{1}.
\par Recently, deep learning techniques have been successfully employed to segment WMH regions \cite{2,3,4,5,6}. For example, Li and colleagues have used U-net \cite{7} in a WMH segmentation challenge \cite{2}. In their proposed approach, they adopted three identical U-nets with different randomly-initialized weights during training and averaged the score maps at the testing stage. U-net has been widely used in segmenting biomedical images because it can work efficiently, when the training samples are few, via data augmentation through elastic deformation \cite{7}. However, the performance of U-net is unstable depending on the randomly initialized weights.
\par For a neural network, its weights are usually randomly initialized so as to break symmetry and make the network coverage faster \cite{8}. The logistic sigmoid activation function has been widely used in various binary classification tasks because it can map any real number to an interval between 0 and 1 \cite{9}. However, Glorot and Bengio found that the logistic sigmoid activation is inappropriate for deep learning networks with random initialization because its mean value may drive the top hidden layer into saturation \cite{10}.
\par Pre-training is usually employed to initialize the weights of a neural network to have a good starting point \cite{11}\cite{12}, and it is helpful for the network to converge to a local optimum during back propagation. However, Pre-training is challenging in medical imaging related classification tasks given that the number of medical images with ground truth labeling is usually very limited, especially for magnetic resonance images (MRIs). Medical image segmentation can be treated as a specific type of binary classification task, with the background being classified as one class (with label value 0) and the region of interest (ROI) being classified as another class (with label value 1).
\par In the work reporting the highest WMH segmentation accuracy, the authors have averaged the outputs of three U-nets to make their model more robust \cite{2}. On such basis, to further reduce the false negative rate, we propose a new post-processing method by thresholding the outputs of the three U-nets followed by averaging. According to the dice similarity coefficient (DSC) analysis, our proposed method can boost the segmentation accuracy of \cite{2} by 1\% .
\par This paper is organized as follows. Section \uppercase\expandafter{\romannumeral2} describes the dataset and the evaluation criteria. Section \uppercase\expandafter{\romannumeral3} details the proposed method. Section \uppercase\expandafter{\romannumeral4} presents our experimental results.  Section \uppercase\expandafter{\romannumeral5} discuss the advantages and potential limitations of the proposed method.
\section{Background}

\subsection{Dataset}
The datasets used in this study came from a WMH segmentation challenge in conjunction with MICCAI 2017 \cite{13}.  There are three datasets acquired from three scanners with each dataset containing 20 subjects. For each subject, there are multi-slice FLAIR and T1 MR images along with ground truth images of WMH regions. Inhomogeneity correction was performed using SPM 12 \cite{14} for all MR images. We performed cross-scanner analysis by using 40 subjects from two scanners as the training data and the other 20 subjects as the testing data. Please note, we used 2D images at the axial direction as our samples rather than the entire 3D images.

\subsection{Evaluation Criteria}
To evaluate the accuracy of an automatically segmented WMH region, we employed the DSC score, which is also known as the similarity index. DSC is a statistic used for comparing the similarity of two sets, which is defined as
    \begin{equation}
    DSC(GS,SEG)=\frac{2 \left| GS \cap SEG \right|}{\left| GS \right| + \left| SEG \right|} ,
    \end{equation}
where $GS$ represents the gold standard segmentation of a WMH region, $SEG$ represents the corresponding automatic segmentation, and $\left| GS \cap SEG \right|$ refers to the overlap region. $\left| \cdot \right|$ represents the sum of the entries of matrix.

\section{Method}
\subsection{Preprocessing and neural network}
To unify the image sizes, we cropped or padded the axial slices of each image to be of size $200 \times 200$. Gaussian normalization was then applied to rescale the voxel intensities of each image. We also conducted data augmentation via rotation, shearing and scaling. The preprocessed FLAIR and T1 images were concatenated to form a tensor, which was served as the input to our neural network. The output was the manually-delineated ground truth segmentation mask.
\par We built a U-net based on the work of Li \cite{2} and Ronneberger \cite{7}. As shown in Fig. 1, the U-net consists of a down-convolutional part (left side) and up-convolutional part (right side). The left side aims at extracting features for classifying each voxel into WMH and non-WMH regions. And the right side aims at locating WMH regions more precisely. The down-convolutional part consists of two $3\times 3$ convolution layers, each followed by a rectified linear unit (ReLU) and a $2 \times 2$ max pooling layer for down-sampling. For the first two convolutional layers, a kernel of size $5 \times 5 $ was used to handle different transformations. Each step in the up-convolutional part involved up-sampling the feature map followed by a $3 \times 3$  convolution layer that reduces the number of the feature channels and two $3 \times 3$ convolutions, each followed by a ReLU. The concatenation was performed between the down-convolutional and up-convolutional parts as shown in Fig. 1 using the gray line. Random initialization was used to initialize the model weights.
    \begin{figure*}[htbp]
    \centering
    \includegraphics[width=6.8in]{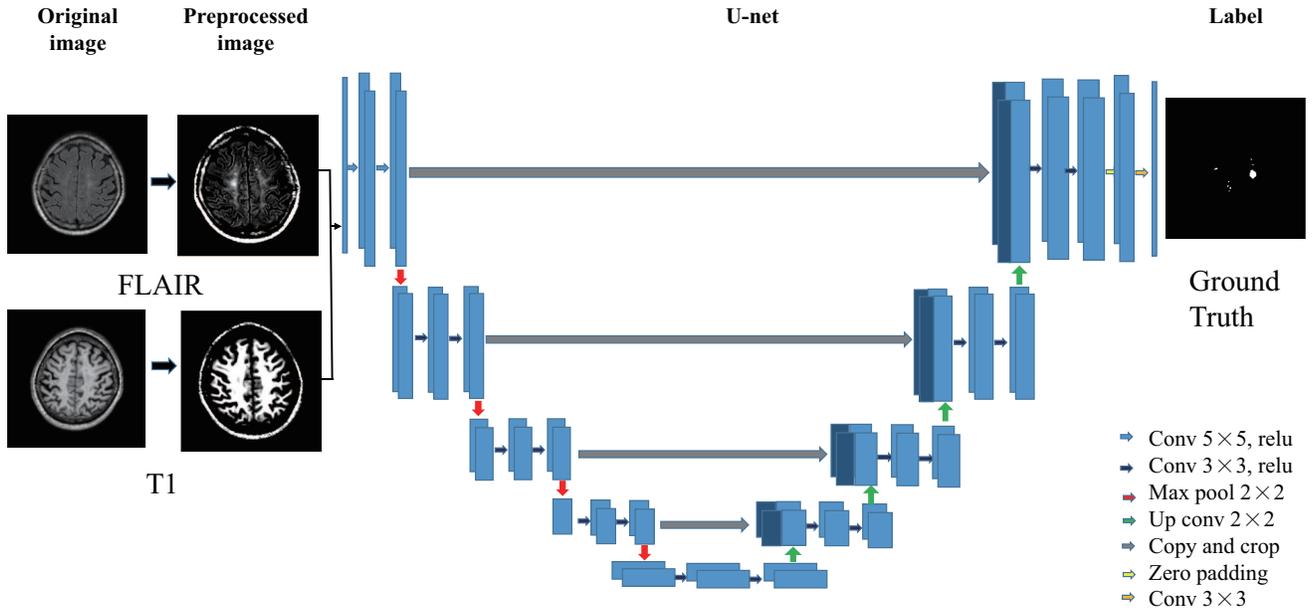}
    \caption{The framwork of the training procedure}
    \label{fig1}
    \end{figure*}

\subsection{Post-processing}
U-net may suffer abnormal local optima because of the random initialization. We assembled the binarization and averaging operations in two different post-processing manners, as shown in Fig. 2. In that figure, panel (a) demonstrates the mean score maps (MSM) method in which the score maps were averaged first and then binarized, and panel (b) shows the mean binary masks (MBM) method in which the score maps were binarized first and then averaged. Both binarization and averaging were performed pixel-wisely. The output of the averaging operation is the mean value of the three input pixel values, and the binarization is conducted as below
    \begin{equation}
    binary\; mask(x,y)= \left\{
        \begin{array}{lr}
        1, &  score\; map(x,y)> 0.5  \\
        0, &  score\; map(x,y)\leq 0.5
        \end{array},
    \right.
    \end{equation}
where $score\; map(x,y)$ denotes the pixel value at the specific location $(x,y)$ in score map, which is the output of a U-net in the testing stage. And $binary\; mask(x,y)$ denotes the pixel value of the binary mask at the corresponding location, which can be treated as the segmentation results.

In Table I, we demonstrate the difference between MSB and MBM. The score values at a specific pixel obtained from the three randomly-initialized U-net 1, U-net 2, and U-net 3 are 0.6, 0.7, 0.1 respectively. The output of the MSM method is 0 whereas the output of the MBM is 1. A possible situation is that both U-net 1 and U-net 2 have converged to a normal optimum whereas U-net 3 suffered an abnormal local optimum. In this case, MBM performed better than MSM because it can reduce the false negative rate caused by the abnormal model which may have been induced by inappropriate randomly-initialized weights.

    \begin{table}[htbp]
    \renewcommand\arraystretch{1.5}
        \centering
        \caption{An Example demonstrating how MSM (Top) and MBM (Bottom) works}

        \begin{tabular}{cccc}   
        \hline
        \textbf{U-net} &\textbf{1} &\textbf{2} &\textbf{3}  \\
        \hline
        \textbf{Score} &0.6 &0.7 &0.1 \\
        \hline
        \textbf{Average} &&0.4667 & \\
        \hline
        \textbf{Threshold} &&0 & \\
        \hline
        \end{tabular}

        \begin{tabular}{cccc}
         \\[3pt]
        \hline
        \textbf{U-net} &\textbf{1} &\textbf{2} &\textbf{3}  \\
        \hline
        \textbf{Score} &0.6 &0.7 &0.1 \\
        \hline
        \textbf{Threshold} &1 &1 &0 \\
        \hline
        \textbf{Average} &&0.6667 & \\
        \hline
        \textbf{Threshold} &&1 & \\
        \hline
        \end{tabular}
    \end{table}

    \begin{figure}[htbp]
        \centering
        \includegraphics[width=5in]{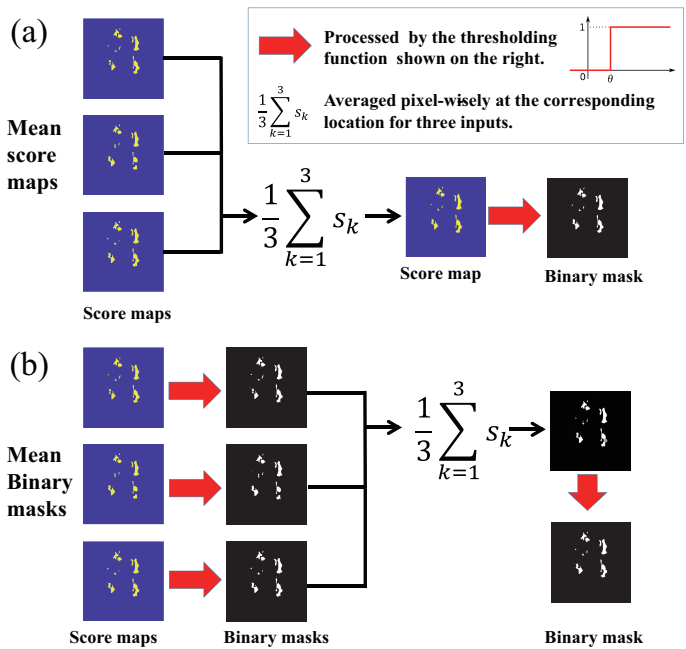}
        \caption{Demonstration of the two post-processing techniques}
        \label{fig2}
    \end{figure}
\subsection{Implementation}
The proposed method was implemented in Python language, using Keras with Tensorflow backend. All experiments were conducted on a Linux machine running Ubuntu 16.04 with 32 GB RAM memory. The U-net training was carried out on a single GTX 1080 Ti with 11 GB RAM memory. We used Adam optimizer with initial learning rate 0.001 and 8 batch size for training.

\section{Results}
\subsection{Visual examinations}

In Fig. 3, we demonstrate the segmentation results for three representative slices. From top to bottom, we respectively show the FLAIR images, the ground truth segmentations overlaid on top of the FLAIR images, the results obtained from the three U-nets (U-net 1, U-net 2, and U-net 3), as well as those obtained from the post-processing methods (MSM and MBM). To better show the segmentation results, selected patches were zoomed by four times and shown on the upper left corners. To clearly reveal the difference between the ground truth segmentation and the segmentation obtained from each automatic method, the ground truth segmentation was shown in red, the automatic segmentation was shown in blue, and their overlapping regions were shown in green. As such, at the lowest five rows (from U-net 1 to MBM), green denotes the overlapping region, red denotes the false negative, and blue denotes the false positive.
\begin{figure}
    \centering
    \includegraphics[width=3.4in]{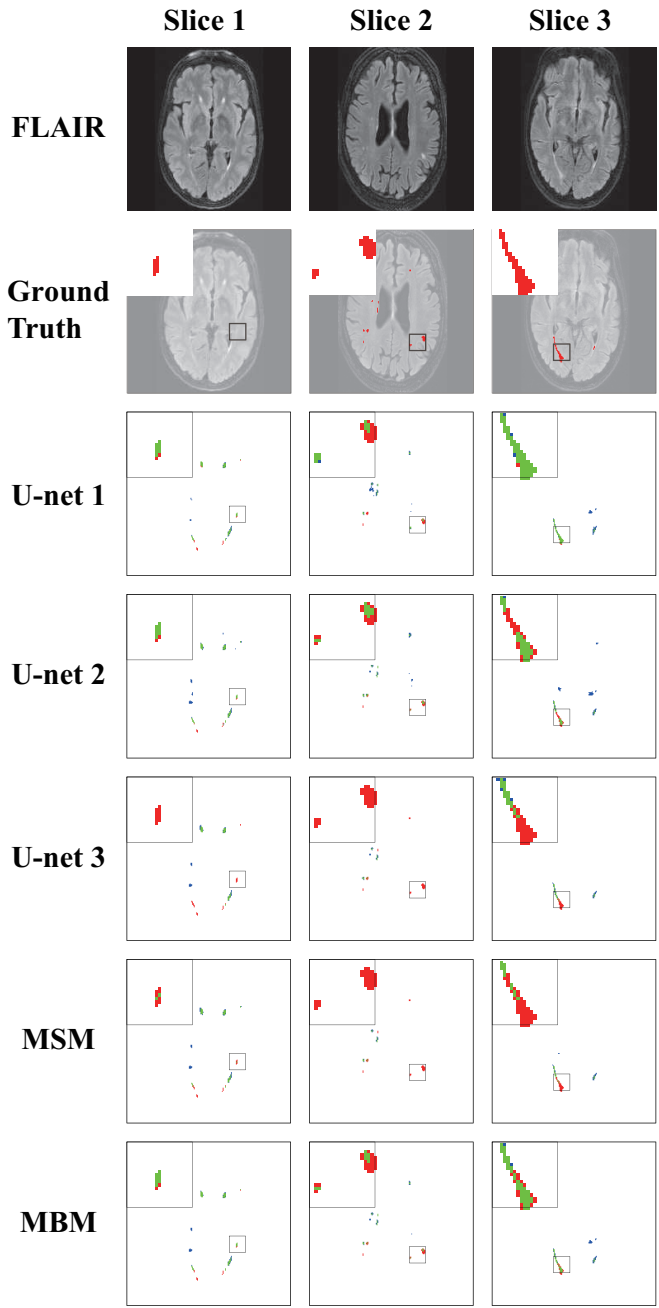}
    \caption{Representative segmentation results from the three U-nets and the two post-processing methods}
    \label{Fig.3}
    \end{figure}

\par From the leftmost column of Fig. 3, we find that the segmentation results from both U-net 1 and U-net 2 are reasonable, whereas that from U-net 3 is far from satisfactory. In such case, MBM performed much better than MSM. This conclusion holds for the other two examples as well (the middle column and the rightmost column).

\subsection{DSC analysis}
Fig. 4 shows the distributions of the DSC scores of the three U-nets, MSM, and MBM, obtained from 20 testing subjects. It is evident that the segmentation performance of U-net 3 is relatively poor. The MBM post-processing method can improve the overall segmentation accuracy, being superior to MSM.
    \begin{figure}
    \centering
    \includegraphics[width=3.4in]{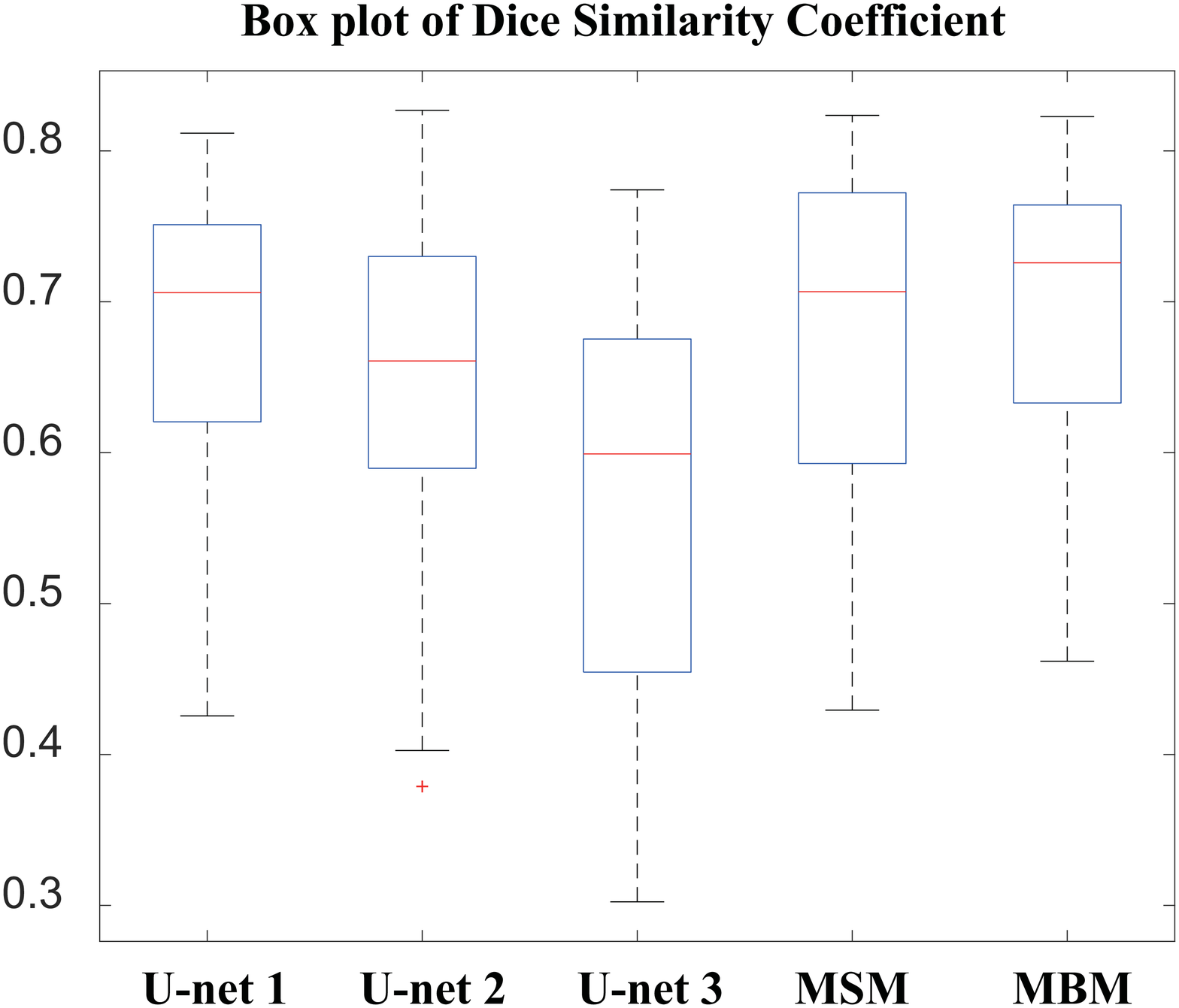}
    \caption{Box plot of DSC for the 20 testing data. (MSM: mean score maps, MSM: mean binary masks)}
    \label{Fig.4}
    \end{figure}
\par In Table II, we tabulate a variety of statistics on the DSC scores of the three U-nets and the two post-processing methods. According to those statistics, we observe that both MSM and MBM are better than each individual U-net and MBM performed even better. With one U-net delivering poor segmentation results, both MSM and MBM can weaken the influence of that poorly-behaving U-net, delivering segmentation results that are superior to those obtained from the best U-net, but MBM is even superior to MSM.
   \begin{table}
    \renewcommand\arraystretch{2}
        \centering
        \caption{Various statistics of the DSC scores for the three U-nets, MBM and MSM}
        \begin{tabular}{lccccc}   %
        \hline
         &\textbf{U-net} 1 &\textbf{U-net 2} &\textbf{U-net 3} & \textbf{MSM} & \textbf{MBM} \\
        \hline
        \textbf{Mean}& 0.6702&	0.6444&	0.5759&	0.6810&	0.6929\\
        \hline
        \textbf{Max} &	0.8118	&0.8269	&0.7742	&0.8235	&0.8228\\
        \hline
        \textbf{Top 75\% }	&0.7510	&0.7300	&0.6753&	0.7722	&0.7641\\
        \hline
        \textbf{Median}	&0.7060&	0.6607	&0.5992	&0.7066	&0.7258\\
        \hline
        \textbf{Top 25\%}	&0.6204	&0.5896	&0.4545 &0.5928&	0.6329\\
        \hline
        \textbf{Min} 	&0.4256	&0.3788	&0.3024	&0.4294	&0.4618\\
       \hline
        \end{tabular}
    \end{table}
\par In clinical application, it is not possible to guarantee that a model will necessarily converge to a good optimum like U-net 1 did. Our post-processing method can reduce the influence of a bad local optimum like U-net 3 has induced and make the model robust.
\section{Discussion and Conclusion}
In this work, we compared two post-processing methods, MSM and MBM, for WMH segmentation in the framework of U-net based learning. We found MBM can further reduce the false negative rate compared to MSM. One potential limitation of this work is that we have not tested the proposed post-processing method in other deep learning models and applications, which will be one of our future endeavors. We believe that the post-processing method can be successfully applied to various deep learning models, especially when pre-training is not available. In addition, there is space for improvement by modifying the structure of the U-net which is the key component in our WMH segmentation approach. Lastly, applying the proposed technique to real-world clinical applications is another future plan of ours.

\section*{Acknowledgment}
We thank the WMH challenge organizer Dr. Hugi J. Kuijf and the joint efforts of the UMC Ultrecht, VU Amsterdam, and NUHS Singapore for making the datasets available for this research. We thank Hongwei Li for his open source code and generous help to rebuild his method.  This work was supported by the Shenzhen Science and Technology Innovation Committee funds (JCYJ20160301113918121), the National Key R\&D Program of China (2017YFC0112404) and the National Natural Science Foundation of China (81501546).

\end{document}